\documentclass[12pt]{article}
\usepackage[final]{neurips2022}
\usepackage{graphicx}
\usepackage{commands}
\usepackage{mathtools}
\usepackage{enumitem}
\usepackage{amsfonts}       
\usepackage{nicefrac}       
\usepackage{microtype}      
\usepackage{outlines}
\usepackage{amsmath}
\usepackage{amssymb}
\usepackage{amsthm}
\usepackage{mathrsfs} 
\usepackage{dsfont}
\usepackage{comment}
\usepackage{subfiles} 

\usepackage{natbib}
\bibpunct{(}{)}{;}{a}{,}{,}
\usepackage{hyperref}
\hypersetup{
colorlinks = true,
urlcolor = blue,
linkcolor = blue,
citecolor = blue,
}

\begin{document}

\title{3DGEN: A GAN-based approach for generating novel 3D models from image data}

\author{Antoine Schnepf
\\	\texttt{a.schnepf@criteo.com} \\
Criteo AI Lab \\
\And
Flavian Vasile	
\\	\texttt{f.vasile@criteo.com} \\
 Criteo AI Lab \\
\And
Ugo Tanielian	
\\	\texttt{u.tanielian@criteo.com}
\\ Criteo AI Lab \\
}

\maketitle
\begin{abstract}
    The recent advances in text and image synthesis show a great promise for the future of generative models in creative fields. However, a less explored area is the one of 3D model generation, with a lot of potential applications to game design, video production, and physical product design. In our paper, we present 3DGEN, a model that leverages the recent work on both Neural Radiance Fields for object reconstruction and GAN-based image generation. We show that the proposed architecture can generate plausible meshes for objects of the same category as the training images and compare the resulting meshes with the state-of-the-art baselines, leading to visible uplifts in generation quality.
\end{abstract}

\section{Introduction}
Generative models such as StableDiffusion \citep{rombach2022high} or DALLE2 \citep{ramesh2022hierarchical} are rapidly changing the boundaries of machine-assisted creativity, especially in the case of image synthesis. 
Researchers and practitioners are inventing new ways to create and remix art, either by text-conditioned image generation, image inpainting or outpainting, full video generation. 
In the same time, the class of Neural Radiance Fields models \citep[NeRF,][]{mildenhall2021nerf}  are making rapid advances in photorealistic 3D model/scene reconstruction from partial views. NeRF uses an implicit volumetric representation to represent 3D scenes, making it possible to render them at an arbitrary resolution with low memory costs. 

Some of the existing work, such as Generative Radiance Fields \citep[GRAF,][]{schwarz2020graf}, has been starting to bridge the gap between reconstruction and generation. GRAF can generate new volumetric models from a set of views of similar objects. However, a major limitation of the GRAF model is that this volumetric representation is not adapted to produce plausible object meshes, and is therefore not a good match for 3D-native creative environments such as game design, virtual-reality (VR) world design, animation. On the other hand, UNISURF \citep{oechsle2021unisurf} showed that radiance fields and implicit surface representations can be unified, and proposed a joint optimization task that both improves NERF and allows to extract 3D meshes.

In this paper, we propose a potential solution for the shortcomings of GRAF, which we name 3DGEN. This solution builds on both GRAF \citep{schwarz2020graf} and UNISURF \citep{oechsle2021unisurf}, and can generate volumetric objects with a corresponding implicit surface, hence making them easily exportable to 3D meshes. In Figure \ref{fig:figure2}, we showcase one potential way to control the object generation: the interpolation in the latent space between two existing object meshes leads to a set of plausible object meshes of the same type (in this case cars and chairs).

\newpage






\section{Our approach}
We begin by introducing a conditional version of UNISURF $g_{\theta}$ which encodes an object conditionally to a shape code and an appearance code, in a disentangled manner.
Similarly to GRAF \citep{schwarz2020graf}, we construct a generator $G_{\theta}$ (sharing parameters with $g_\theta$) by stacking (i) a module that casts rays, (ii) $g_{\theta}$ that conditionally computes the emitted radiance and occupancy probabilities along the cast rays, and (iii) a differentiable volumetric renderer to produce the output images \citep{mildenhall2021nerf}. We introduce the discriminator $D_{\phi}$, a convolutionnal neural network. 

We train this setup with the non-saturating GAN objective \citep{goodfellow2014GAN} with R1-regularization \citep{WhichGanMethodsConverge}, to which a smoothing term for the implicit surface is added \citep{oechsle2021unisurf} (more details on each term can be found in Appendix):
\begin{equation}
    \min_{\theta} \max_{\phi}  \Bigl( \mathcal{L}_{\text{adv}}(\theta, \phi) - \lambda \mathcal{R}_1(\phi) + \gamma \mathcal{L}_{\text{smooth}}(\theta)  \Bigr)
\end{equation}

\paragraph{Implementation details and surface extraction.}
Our model is initialized such that the initial implicit surfaces are spheres (\cite{oechsle2021unisurf}, \cite{2020GeometricRegularization}). As the training progresses, the points are sampled along rays within a narrowing interval centered around the first intersection with the implicit surface \citep{oechsle2021unisurf}. By doing so, in the early stages of the training, the formulation is similar to GRAF, while in the later stages of the training, points are sampled close to the implicit surface. It is therefore possible to extract a well defined surface with the Marching Cube algorithm \citep{lorensen1987marching}.

\begin{figure}
    \centering
    \includegraphics[width=1\textwidth]{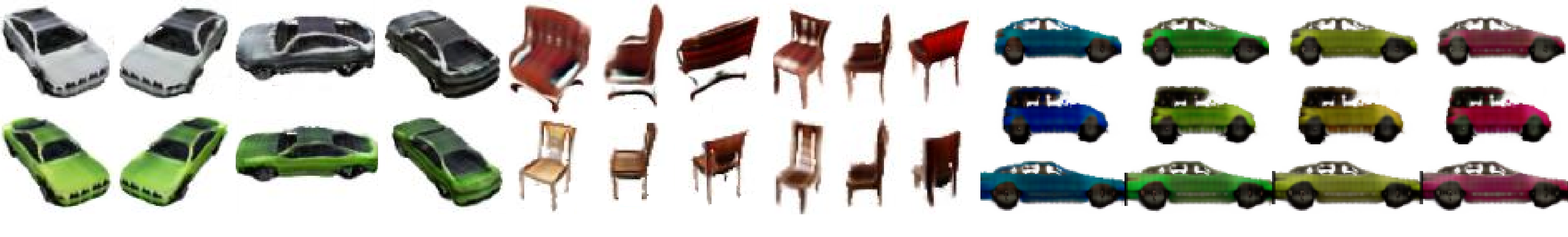}  
    \caption{Left: 3DGEN with cars and chairs. Right: disentanglement of shape and appearance.}
    \label{fig:figure1}
\end{figure}

\textbf{Experiments.} We test our model on (i) a dataset of cars rendered from The Carla Driving simulator \citep{carladataset} and (ii) a dataset of chairs, rendered from Photoshapes \citep{photoscape}. Figure \ref{fig:figure1} shows the generated cars and chairs, as well as disentanglement of shape and appearance. Figure \ref{fig:figure2} shows latent space interpolation and the exportation to meshes. In the Appendix, 3DGEN is compared to the baseline GRAF for camera poses interpolations in Figure \ref{fig:Interpolation_cars_and_chairs} and meshes extraction in Figure \ref{fig:meshes}. To evaluate both methods, we report the Frechet Inception Distance \citep[FID, ][]{heusel2017gans}. GRAF / Ours:  $71 / 97$ (Cars); $48 / 126$ (Chairs). For fairness, both methods are trained for the same duration on similar hardware.

\begin{figure}
    \centering
    \includegraphics[width=0.558\textwidth]{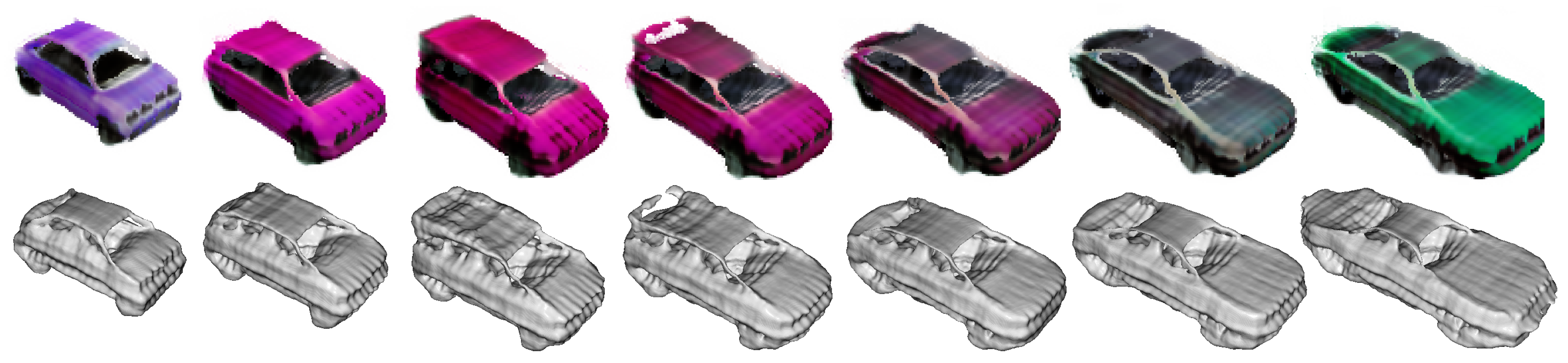}  
    \includegraphics[width=0.341\textwidth]{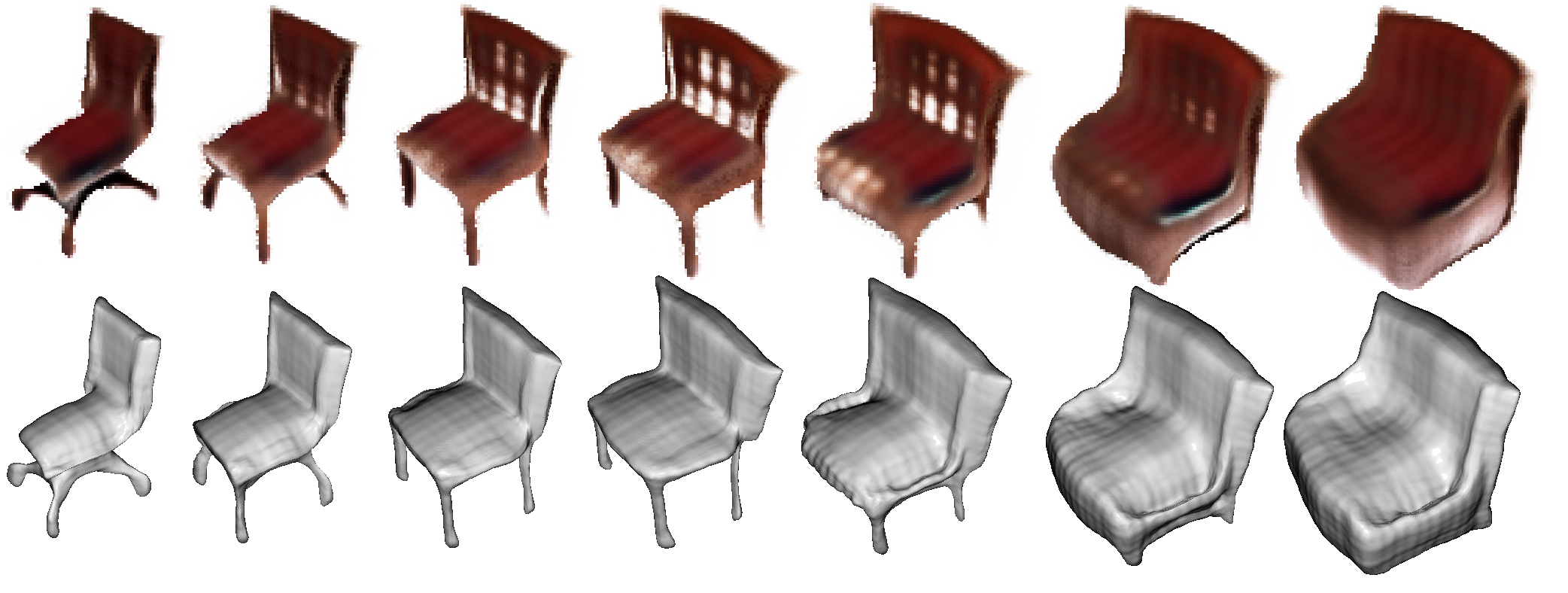} 
    \caption{Latent space interpolation and the corresponding meshes.}
    \label{fig:figure2}
\end{figure}

\section{Conclusion and future work}
This work present 3DGEN, a generative model that unifies radiance fields and implicit surfaces. The model can learn an underlying distribution of radiance fields and surfaces from a dataset composed only of 2D images of objects of a similar class, and therefore \emph{generate new objects} from this class. During inference, the model can both render views from any angle and easily export to a mesh based representation, which makes it applicable to 3D content creation. In our future work we intend to further improve the quality of the generated objects by reducing artefacts in the shapes and producing more diverse outputs. The ability to export to colored meshes would also be an useful improvement.



\newpage

\section{Appendix}

\subsection{Background}
This section summarizes the mathematical formalism required for our proposed model 3DGEN. 

\vspace{0.6cm}

\textbf{Neural Radiance Field} \citep{mildenhall2021nerf}.
Neural radiance fields are neural networks parameterized by $\theta \in \Theta$ mapping a spatial location $\mathbf{x} \in \mathds{R}^3$ and a viewing direction $\mathbf{d} \in \mathds{R}^3$ to a view-dependant color $c_{\theta}(\mathbf{x}, \mathbf{d})$ and a volumetric opacity $\sigma_{\theta}(\mathbf{x}) \in \mathds{R}^+$:
\begin{equation}
    f_{\theta} : \mathbf{x}, \mathbf{d} \longrightarrow c_{\theta}(\mathbf{x}, \mathbf{d}), \sigma_{\theta}(\mathbf{x})
\end{equation}
A neural radiance field encodes a 3D scene or object in a volumetric representation.

\vspace{0.6cm}

\textbf{Differentiable volume rendering.}  Given the the camera origin $\mathbf{o} \in \mathds{R}^3$ and a viewing direction $\mathbf{d}$, a ray $\mathbf{r}=\{\mathbf{o} + t\mathbf{d} | t \in \mathbb{R}^+\}$ is cast through the radiance field.
Given N points $\{\mathbf{x}_i\}$ sampled along this ray, differentiable volume rendering approximates the perceived color of the ray as follows:
\begin{equation}\label{eq:volume rendering}
    \hat{C}(\mathbf{r}) = \sum^N_{i=1}T_i \alpha_i c_\theta(\mathbf{x}_i,\mathbf{d})
\end{equation}
\begin{equation}\label{eq:alpha_i}
    \alpha_i = 1 - \exp(-\sigma_\theta(\mathbf{x}_i)\delta_i))
\end{equation}
\begin{equation}
    T_i = \prod_{j<i} (1 - \alpha_j)
\end{equation}
where $T_i$ is the accumulated transmittance along the ray $\mathbf{r}$ and $\delta_i = ||\mathbf{x}_{i+1} - \mathbf{x}_i||_2$ is the distance between two adjacent points.

By casting $K \times K$ carefully chosen rays and rendering their respective colors, an image of the encoded object is obtained, where K is the desired resolution. 

\vspace{0.6cm}

\textbf{UNISURF} \citep{oechsle2021unisurf}. Assuming solid non-transparent objects, i.e. $\sigma_{\theta}(\mathbf{x}_i) \in \{0, +\infty \}$,  $\alpha_i$ can be reinterpreted as the occupancy probability at position $\mathbf{x}_i$ from equation \ref{eq:alpha_i}.
We can therefore derive an implicit surface $\mathcal{S}_{\theta}$:

\begin{equation}
    \mathcal{S}_{\theta} = \{\mathbf{x} \in \mathbb{R}^3 | \alpha_{\theta}(\mathbf{x}) = 0.5 \}
\end{equation}

A regularization loss $\mathcal{L}_{\text{smooth}}$ on the implicit surface is introduced: 

\begin{equation}
    \mathcal{L}_{\text{smooth}}(\theta) = \sum_{\mathbf{x} \in \mathcal{S}_{\theta}}\left\|\mathbf{n}_{\theta}\left(\mathbf{x}\right)-\mathbf{n}_{\theta}\left(\mathbf{x}+\boldsymbol{\epsilon}\right)\right\|_2
\end{equation}

Here $\boldsymbol{\epsilon}$ is a small perturbation and $\mathbf{n}(\mathbf{x}_s)$ denotes the surface normal at position $\mathbf{x}_s$. The surface normal can be computed using the formula:

\begin{equation} \label{eq:surface normal}
\mathbf{n_{\theta}}(\mathbf{x})=\frac{ \nabla \alpha_\theta(\mathbf{x})}{\|\nabla \alpha_\theta (\mathbf{x})\|_2}
\end{equation}

We adapt this loss for adversarial training: 

\begin{equation}
    \mathcal{L}_{\text{smooth}}(\theta) = 
    \mathbb{E}_{\mathbf{z}_s \sim p_{\text{latent}}, \boldsymbol{\xi} \sim p_\text{cam}} \biggl[ \ 
    \sum_{\mathbf{x} \in \mathcal{S}_{\theta}(\mathbf{z}_s) \cap \mathcal{R}(\boldsymbol{\xi})}\left\|\mathbf{n}_{\theta}\left(\mathbf{x}, \mathbf{z}_s\right)-\mathbf{n}_{\theta}\left(\mathbf{x}+\boldsymbol{\epsilon}, \mathbf{z}_s\right)\right\|_2
    \biggr]
\end{equation}

where the implicit surface $\mathcal{S}_\theta(\mathbf{z}_s)$ and its normal $\mathbf{n}_{\theta}(\mathbf{x}, \mathbf{z}_s)$ are now conditional to a shape code $\mathbf{z}_s$. $\mathcal{R}(\boldsymbol{\xi})$ denotes the set of rays cast in the radiance field under the camera parameters $\boldsymbol{\xi}$.

\textbf{GRAF} \citep{schwarz2020graf}.
Built upon the original NeRF formulation, a conditional neural radiance field (cNeRF) generates radiance fields conditionally to an appearance code $z_a$ and a shape code $z_s$: 

\begin{equation}
g_{\theta} : \mathbf{x}, \mathbf{d}, \mathbf{z_s}, \mathbf{z_a} \longrightarrow c_{\theta}(\mathbf{x}, \mathbf{d}, \mathbf{z_s}, \mathbf{z_a}), \sigma_{\theta}(\mathbf{x}, \mathbf{z_s})
\end{equation}

The generator $G_\theta$ is composed of three modules: 

\begin{enumerate}[label=(\roman*)]
    \item a module that sample the camera parameters $\boldsymbol{\xi}$ and subsequently cast $K \times K$ rays
    \item a cNeRF $g_{\theta}$ to conditionally compute radiances and volumetric opacities along the sampled rays
    \item a volume renderer to produce the output image
\end{enumerate}

To train the discriminator $D_{\phi}$ and the generator $G_{\theta}$, the adversarial loss $\mathcal{L}_{\text{adv}}(\theta, \phi)$ and the regularization term $\mathcal{R}_1(\phi)$ are introduced: 

\begin{multline}
\mathcal{L}_{\text{adv}}(\theta, \phi) = 
\mathbb{E}_{\mathbf{I} \sim p_{\text {data }}, \boldsymbol{\nu} \sim p_{\text{patch}}}
\Bigl[f\left(-D_{\phi}(\Gamma(\mathbf{I}, \boldsymbol{\nu}))\right)\Bigr] 
\\ +  
\mathbb{E}_{\mathbf{z_s}, \mathbf{z_a}\sim p_{\text {latent }}, \boldsymbol{\xi} \sim p_{\text{cam}}, \boldsymbol{\nu} \sim p_{\text{patch}}}\Bigl[f\left(D_{\phi}(G_{\theta}(\mathbf{z_s}, \mathbf{z_a}, \boldsymbol{\xi}, \boldsymbol{\nu} ))\right)\Bigr] 
\end{multline}

\begin{equation}
\mathcal{R}_1(\phi)  = \mathbb{E}_{\mathbf{I} \sim p_{\text {data }}, \boldsymbol{\nu} \sim p_{\text{patch}}}
\Bigl[\bigl|\bigl|\nabla D_{\phi}\left(\Gamma(\mathbf{I}, \boldsymbol{\nu})\right)\bigr|\bigr|_2^2\Bigr] 
\end{equation}

with $f(x) = -\log(1 + \exp(-x))$. Here, $\boldsymbol{\nu}$ denotes denotes the patching strategy and $\Gamma$ the patching operator to transform the training images into $K \times K$ patches. Note that during inference, the generator $G_\theta$ can be used to generate images at any resolution. The patch constraint only holds during training, to make this setup trainable in practice. For more details, we refer the reader to \cite{schwarz2020graf}.

\newpage

\subsection{Qualitative comparison between 3DGEN and GRAF}

\vspace{2cm}
\begin{figure}[h]
    \centering
    \includegraphics[width=0.45\textwidth]{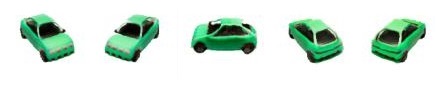}  
    \includegraphics[width=0.45\textwidth]{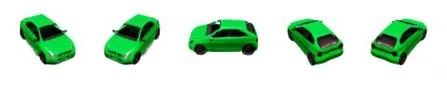}  
    \includegraphics[width=0.45\textwidth]{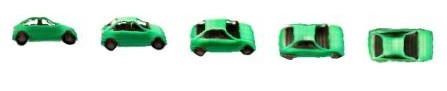} 
    \includegraphics[width=0.45\textwidth]{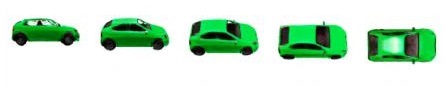}  
    
    \vspace{0.1cm}
    
    \includegraphics[width=0.45\linewidth]{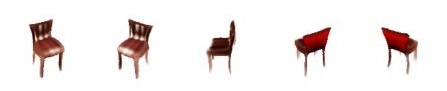} 
    \includegraphics[width=0.45\linewidth]{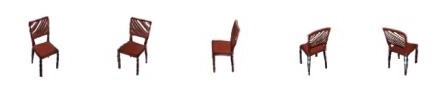}  
    \includegraphics[width=0.45\linewidth]{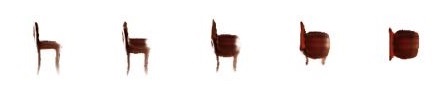}
    \includegraphics[width=0.45\linewidth]{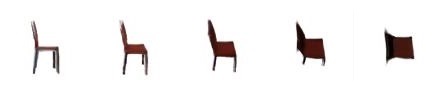}

    \caption{Camera poses interpolations on both Cars and Chairs. 3DGEN (left) and GRAF (right).}
    \label{fig:Interpolation_cars_and_chairs}
    
\end{figure}

\vspace{3cm}
\vspace{1cm}
\begin{figure}[h]
    \centering
    \includegraphics[width=0.9\linewidth]{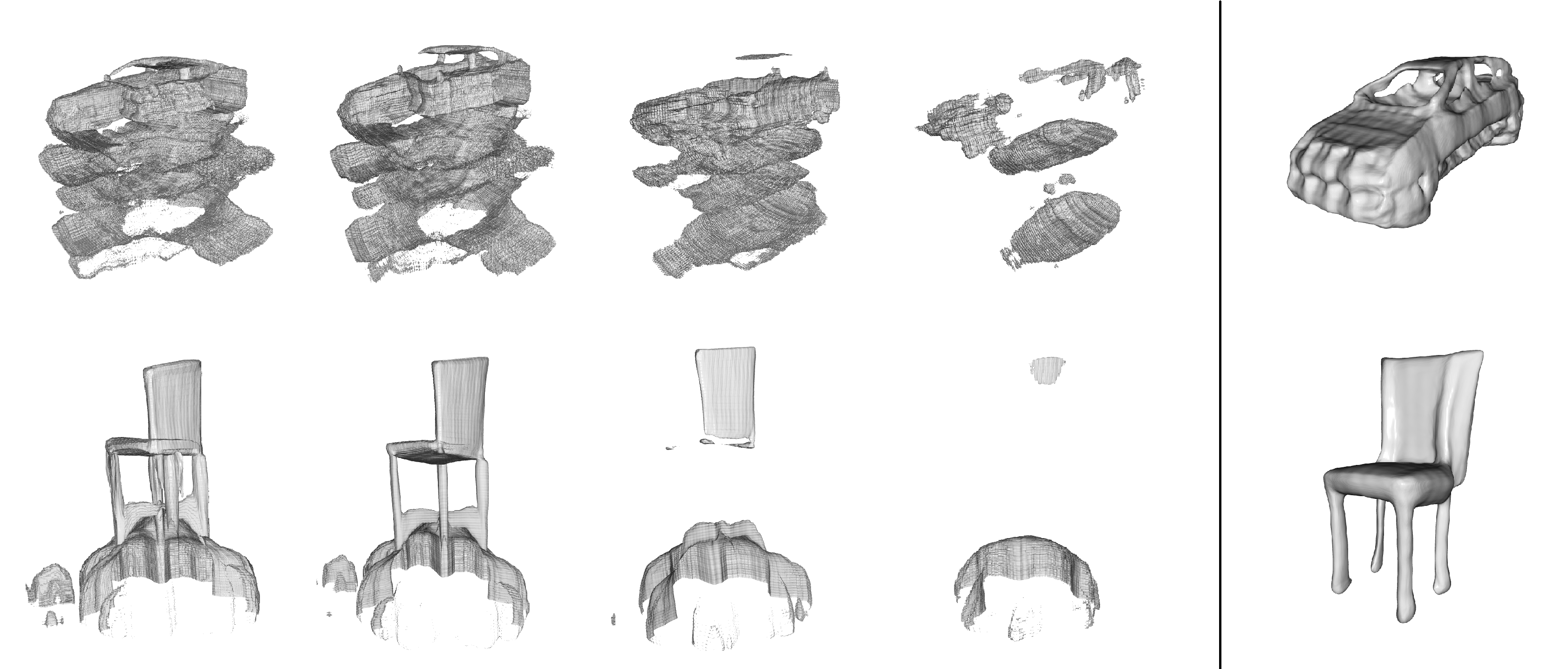}  

    \caption{Extracting surface meshes from GRAF at level set $\sigma \in \{1, 10, 50, 100\}$ (left) and from 3DGEN (right).}
    \label{fig:meshes}
\end{figure}

\end{document}